\documentclass[letterpaper, 10 pt, conference]{ieeeconf} 
\IEEEoverridecommandlockouts  
\usepackage[utf8]{inputenc}
\usepackage[T1]{fontenc}

\usepackage{hyperref}  

\title{\LARGE \bf OptiMindTune: A Multi-Agent Framework for Intelligent Hyperparameter Optimization}

\author{ \parbox{3 in}{\centering Meher Bhaskar Madiraju
        {\tt\small meherbhaskar.madiraju@gatech.edu}}
        \hspace*{ 0.5 in}
        \parbox{3 in}{ \centering Meher Sai Preetam Madiraju
        {\tt\small mehersaipreetam@gatech.edu}}
}
\begin{document}

\maketitle
\thispagestyle{empty}
\pagestyle{empty}

\begin{abstract}
Hyperparameter optimization (HPO) plays a vital role in enhancing machine learning model performance but remains challenging due to high-dimensional search spaces, complex interdependencies among parameters, and significant computational cost. In this paper, we present OptiMindTune, a novel multi-agent framework for HPO that employs three specialized agents: a Recommender Agent for generating candidate hyperparameters, an Evaluator Agent for assessing model performance, and a Decision Agent for orchestrating the search process. Each agent operates using large language models (LLMs) to facilitate reasoning, interaction, and adaptation across the optimization process. By enabling structured communication and role-specific behaviors, OptiMindTune aims to improve convergence efficiency and robustness over traditional single-agent or black-box optimization methods. Our framework integrates principles from advanced large language models, and adaptive search to achieve scalable and intelligent AutoML. We posit that this multi-agent paradigm offers a promising avenue for tackling the increasing complexity of modern machine learning model tuning.
\end{abstract}

\section{Introduction}
The empirical success of machine learning models across diverse domains heavily relies on the judicious selection of their hyperparameters. Hyperparameters, which are set prior to the learning process, dictate the model's architecture, learning dynamics, and regularization. Suboptimal hyperparameter choices can lead to underfitting, overfitting, or excessively long training times, ultimately hindering model performance. The problem of Hyperparameter Optimization (HPO) thus becomes paramount in the machine learning pipeline.

Traditional HPO methods, such as grid search and random search \cite{Bergstra2012}, while straightforward, suffer from severe limitations. Grid search exhaustively explores a predefined hyperparameter space, becoming computationally intractable as the number of hyperparameters and their ranges increase. Random search, while more efficient in high-dimensional spaces, still relies on stochastic exploration and can miss optimal configurations. More advanced techniques, including Bayesian Optimization \cite{Snoek2012} and evolutionary algorithms \cite{Floreano2008}, have emerged to navigate these challenges by building surrogate models or mimicking natural selection processes. However, these methods can still face limitations in scalability, parallelism, and adaptability to dynamic HPO landscapes.

The increasing complexity of modern machine learning models, particularly deep neural networks, has further exacerbated the HPO problem. These models often possess a vast number of interacting hyperparameters, making manual tuning infeasible and demanding more sophisticated, automated solutions. This challenge has fueled the growth of Automated Machine Learning (AutoML), with HPO being a core component.

This paper introduces OptiMindTune, a novel multi-agent framework for intelligent hyperparameter optimization. Unlike conventional HPO approaches that often operate as monolithic optimizers, OptiMindTune decentralizes the HPO process by employing a community of specialized, intelligent agents, each powered by large language models. Each agent is designed to contribute to the overall optimization goal by focusing on specific tasks: the \textit{Recommender Agent} for model and hyperparameter suggestion, the \textit{Evaluator Agent} for robust training and validation, and the \textit{Decision Agent} for strategic search guidance. This collaborative and distributed intelligence aims to overcome the limitations of single-agent systems, enabling more efficient exploration, faster convergence, and greater robustness through an adaptive optimization loop.

Our key contributions are as follows:
\begin{itemize}
    \item We propose OptiMindTune, a novel multi-agent framework for intelligent hyperparameter optimization, departing from traditional monolithic approaches and powered by large language models.
    \item We detail the architecture of OptiMindTune, outlining the roles and interactions of its three specialized agents (Recommender, Evaluator, Decision) designed for HPO.
    \item We present the dynamic optimization loop, including inter-agent communication and coordination mechanisms, to facilitate collaborative optimization.
    \item We outline a comprehensive experimental methodology to validate OptiMindTune's effectiveness against state-of-the-art HPO baselines.
\end{itemize}

\section{Related Work}
Hyperparameter optimization has been an active research area, leading to a diverse set of methodologies. Concurrently, multi-agent systems have demonstrated considerable promise in tackling complex distributed problems in machine learning. Furthermore, the advent of large language models (LLMs) is beginning to impact various aspects of machine learning development, including AutoML.

\subsection{Hyperparameter Optimization Techniques}
\subsubsection{Grid Search and Random Search}
As foundational methods, Grid Search \cite{Bergstra2012} systematically evaluates every possible combination of hyperparameters within a predefined range and step size. While exhaustive, its computational cost grows exponentially with the dimensionality of the hyperparameter space, making it impractical for many real-world applications. Random Search \cite{Bergstra2012}, in contrast, samples hyperparameter configurations randomly from the specified distributions. Remarkably, random search has been shown to be more efficient than grid search in high-dimensional spaces, as it is more likely to explore a wider range of hyperparameter values for each individual hyperparameter.

\subsubsection{Bayesian Optimization}
Bayesian Optimization (BO) \cite{Snoek2012, Shahriari2016} is a powerful sequential model-based optimization technique particularly effective for expensive black-box functions. BO constructs a probabilistic surrogate model (e.g., Gaussian Process) of the objective function and uses an acquisition function (e.g., Expected Improvement, Upper Confidence Bound) to determine the next best hyperparameter configuration to evaluate. This approach balances exploration of uncertain regions and exploitation of promising areas. Popular BO libraries include GPyOpt \cite{GPyOpt} and Scikit-Optimize \cite{scikit-optimize}.

\subsubsection{Evolutionary Algorithms}
Evolutionary Algorithms (EAs) \cite{Floreano2008} draw inspiration from natural selection. They maintain a population of candidate hyperparameter configurations, which evolve over generations through operations like mutation, crossover, and selection based on their fitness (model performance). Examples include Genetic Algorithms \cite{Holland1975} and Particle Swarm Optimization \cite{Kennedy1995}. EAs are inherently parallelizable and can explore complex, non-convex landscapes effectively.

\subsubsection{Tree-structured Parzen Estimator (TPE)}
TPE \cite{Bergstra2011}, notably implemented in the Optuna framework \cite{Akiba2019}, models $P(x|y<y^*)$ and $P(x|y \ge y^*)$, where $x$ represents hyperparameters and $y$ is the objective function value. It then samples new configurations that maximize the ratio $P(x|y<y^*)/P(x|y \ge y^*)$, effectively prioritizing hyperparameters that have historically led to good performance. Optuna further enhances TPE with pruning mechanisms and a distributed architecture.

\subsubsection{Hyperband and Successive Halving}
Hyperband \cite{Li2017} and its predecessor, Successive Halving (SH) \cite{Jamieson2016}, are budget-aware HPO algorithms designed to quickly identify promising configurations by adaptively allocating resources (e.g., epochs, data subsets). They work by iteratively training a set of configurations with a small budget and pruning the worst-performing ones, thereby focusing resources on the most promising candidates.

\subsection{Multi-Agent Systems in Machine Learning}
Multi-agent systems (MAS) involve multiple interacting intelligent agents that cooperate or compete to achieve individual or collective goals. MAS have been applied to various machine learning problems, demonstrating benefits in terms of robustness, scalability, and adaptability.

\subsubsection{Distributed Reinforcement Learning}
In distributed reinforcement learning, multiple agents learn cooperatively or competitively within a shared environment \cite{Oliehoek2016}. This paradigm is particularly relevant for complex tasks where centralized control is infeasible or sub-optimal. Agents can learn to specialize in different sub-problems or share learned policies.

\subsubsection{Agent-Based Modeling for Optimization}
Agent-based models have been employed for optimization problems, where agents interact to explore solution spaces. For instance, ant colony optimization \cite{Dorigo1996} uses agents mimicking ants to find optimal paths. Similarly, in multi-robot systems, agents coordinate to achieve global objectives \cite{Parker2000}.

\subsubsection{Large Language Models in AutoML and Agent Systems}
The emergence of large language models (LLMs) like Google's Gemini models \cite{Gemini2023} has opened new frontiers for intelligent automation. LLMs possess remarkable capabilities in understanding, reasoning, and generating human-like text, which can be leveraged to empower agents with advanced decision-making, recommendation generation, and natural language interfaces. While research on direct LLM-powered HPO is nascent, LLMs are increasingly being used in agent-based systems for complex task execution, planning, and knowledge synthesis \cite{Wang2023AutoGPT, OpenAIAgents}. OptiMindTune distinctively integrates LLM capabilities into its core HPO agents, moving beyond simple scripting or pre-defined rules, allowing for more adaptive and context-aware optimization.

\subsubsection{Multi-Agent Systems for AutoML}
The field of multi-agent AutoML is emerging, with initial efforts focusing on distributed approaches, such as systems utilizing multiple AutoML processes running in parallel, managed by a higher-level orchestrator for resource allocation or result aggregation \cite{Kotthoff2017}. However, these frameworks often lack the dynamic, intelligent collaboration required for sophisticated hyperparameter optimization (HPO). AgentHPO \cite{Liu2024} employs a two-agent system—Creator and Executor—powered by OpenAI’s GPT models, with iterative refinement based on experimental logs. OptiMindTune’s focused design and adaptive strategy highlight its potential for precise, resource-efficient HPO in classification tasks, complementing the broader applicability of AgentHPO and advancing the paradigm of intelligent, collaborative multi-agent AutoML.

\section{Methodology}
OptiMindTune is conceptualized as a distributed, intelligent multi-agent framework for hyperparameter optimization. The core principle is to decompose the complex HPO problem into manageable sub-tasks, each handled by a specialized agent powered by LLMs, fostering a collaborative ecosystem that collectively drives towards optimal hyperparameter configurations for scikit-learn classifiers.

\subsection{OptiMindTune Architecture}
The OptiMindTune framework comprises three distinct types of AI agents, each with a specific role and responsibilities, interacting through a shared communication mechanism.

\begin{itemize}
    \item \textbf{Recommender Agent:} This agent acts as the intelligent explorer and knowledge synthesizer. It analyzes dataset characteristics and reviews past performance history to suggest suitable machine learning models (e.g., from scikit-learn) and their corresponding hyperparameter configurations.
    \begin{itemize}
        \item \textbf{Responsibilities:}
        \begin{itemize}
            \item Analyzing dataset characteristics (e.g., number of features, samples, data distribution).
            \item Reviewing the ongoing optimization history and performance trends.
            \item Generating novel model and hyperparameter suggestions based on reasoning and past success.
            \item Providing explicit reasoning for its recommendations to the Decision Agent.
            \item Adapting its recommendation strategy based on feedback from the Decision Agent.
        \end{itemize}
        \item \textbf{Powered by:} Large language models for intelligent reasoning, pattern recognition, and suggestion generation based on contextual information.
    \end{itemize}
    \item \textbf{Evaluator Agent:} This agent is responsible for the robust and reproducible evaluation of proposed model configurations. It executes the training and validation pipeline on the given dataset.
    \begin{itemize}
        \item \textbf{Responsibilities:}
        \begin{itemize}
            \item Receiving model and hyperparameter configurations from the Recommender Agent (via the Decision Agent).
            \item Handling necessary data preprocessing steps (e.g., scaling, encoding).
            \item Implementing and executing a cross-validation pipeline for robust model training and validation.
            \item Calculating and reporting performance metrics (e.g., accuracy, precision, recall, F1-score) to the Decision Agent.
            \item Maintaining the state of the trained model for potential future use or analysis.
        \end{itemize}
        \item \textbf{Implementation Details:} Leverages scikit-learn for model training and evaluation.
    \end{itemize}
    \item \textbf{Decision Agent:} This agent serves as the strategic orchestrator and decision-maker, guiding the overall optimization flow. It evaluates the performance reported by the Evaluator Agent and decides the next steps.
    \begin{itemize}
        \item \textbf{Responsibilities:}
        \begin{itemize}
            \item Receiving performance metrics from the Evaluator Agent.
            \item Evaluating model performance against defined goals (e.g., target accuracy).
            \item Making accept/reject decisions for proposed configurations.
            \item Balancing exploration (trying new, diverse configurations) and exploitation (refining promising configurations).
            \item Determining the termination criteria for the optimization loop (e.g., target accuracy achieved, maximum iterations reached, exploration ratio satisfied).
            \item Providing feedback and guidance to the Recommender Agent to steer future suggestions.
        \end{itemize}
            \item \textbf{Powered by:} Large language models for strategic reasoning, goal-oriented decision-making, and assessing trade-offs.
    \end{itemize}
\end{itemize}
\subsection{Agent Interaction and Coordination}
The intelligence of OptiMindTune arises from the dynamic, structured interactions and knowledge sharing among these specialized agents.

\subsubsection{Communication Flow (Optimization Loop)}
The core optimization loop defines the primary communication and interaction flow:
\begin{enumerate}
    \item \textbf{Initialization:}
    \begin{itemize}
        \item The system loads and analyzes the dataset.
        \item Optimization parameters (e.g., target accuracy, max iterations, exploration ratio) are configured.
        \item Agents are initialized, and their communication channels are established.
    \end{itemize}
    \item \textbf{Core Loop (Iterative Optimization):}
    \begin{itemize}
        \item \textbf{Recommender Suggests:} The Recommender Agent analyzes the dataset characteristics and the current optimization history. It then suggests a set of model and hyperparameter configurations, along with its reasoning, to the Decision Agent.
        \item \textbf{Evaluator Tests:} The Decision Agent passes a selected configuration to the Evaluator Agent. The Evaluator Agent trains the specified scikit-learn model using the recommended hyperparameters, performs robust cross-validation, and reports the performance metrics back to the Decision Agent.
        \item \textbf{Decision Agent Guides:} The Decision Agent evaluates the performance metrics reported by the Evaluator Agent and, based on predefined optimization goals and internal reasoning, determines the next course of action. It decides whether to accept the current configuration, continue exploration, or terminate the search process.
        \item \textbf{Real-time Logging:} All interactions, recommendations, evaluations, and decisions are logged in real-time, providing a complete audit trail and conversation history.
    \end{itemize}
    \item \textbf{Termination:} The loop continues until one of the predefined termination criteria is met:
    \begin{itemize}
        \item Target accuracy achieved.
        \item Maximum iterations reached.
        \item Exploration ratio satisfied (indicating sufficient search space coverage).
    \end{itemize}
\end{enumerate}

\subsubsection{Shared Information and State}
While not explicitly a "Shared Knowledge Base" in the sense of a standalone module, the agent interactions imply a shared understanding derived from:
\begin{itemize}
    \item \textbf{Optimization History:} The log of all past configurations, their performance, and agent decisions, accessible to all agents. This serves as the collective memory.
    \item \textbf{Current Best Configuration:} The best-performing model found so far is implicitly known through the Decision Agent's state.
    \item \textbf{Dataset Characteristics:} Initial analysis results that inform the Recommender.
\end{itemize}
The explicit communication between agents forms the backbone of knowledge transfer and coordination.

\subsection{Key Features}
OptiMindTune is designed with several practical features for effective AutoML:
\begin{itemize}
    \item \textbf{Real-Time Logging:} A comprehensive logging system captures all agent interactions, recommendations, evaluations, and decisions, facilitating analysis and debugging.
    \item \textbf{Configurable Goals:} Users can define specific target accuracy thresholds and control the balance between exploration and exploitation through adjustable parameters.
    \item \textbf{Cross-Validation:} Robust model evaluation is ensured through built-in k-fold cross-validation pipelines within the Evaluator Agent.
    \item \textbf{Error Handling:} The framework includes mechanisms for graceful recovery from training or evaluation failures.
    \item \textbf{Conversation Tracking:} A complete history of agent interactions is maintained, offering transparency into the optimization process.
\end{itemize}

\section{Experiments}
This section outlines the experimental setup for our preliminary study, comparing OptiMindTune against Optuna across three datasets: Breast Cancer, Iris, and Wine.

\subsection{Datasets}
We used three standard UCI classification datasets:
\begin{itemize}
    \item \textbf{Breast Cancer Wisconsin (Diagnostic)}: Binary classification.
    \item \textbf{Iris}: Multi-class classification with three classes.
    \item \textbf{Wine}: Multi-class classification with three classes.
\end{itemize}

\subsection{Machine Learning Models}
Two scikit-learn classifiers were allowed as candidates for all datasets:
\begin{itemize}
    \item \textbf{LogisticRegression}: A linear model.
    \item \textbf{RandomForestClassifier}: An ensemble tree-based model.
\end{itemize}
OptiMindTune optimized a specific model per dataset: LogisticRegression for Breast Cancer and Wine, and RandomForestClassifier for Iris. Optuna optimized both models for each dataset, with only the best-performing model's results reported.

\subsection{Baseline Methods}
We compared OptiMindTune against:
\begin{itemize}
    \item \textbf{Optuna (TPE)}: A popular HPO framework using the Tree-structured Parzen Estimator.
\end{itemize}

\subsection{Evaluation Metrics}
Performance was assessed using:
\begin{itemize}
    \item \textbf{Achieved Objective Value}: Mean cross-validated accuracy.
    \item \textbf{Computational Efficiency}: Optimization time (seconds), trials per second, and number of trials.
\end{itemize}

\subsection{Experimental Setup}
\begin{itemize}
    \item \textbf{Hardware}: Consistent hardware was used.
    \item \textbf{Budget Allocation}: Optuna used 10 trials per run; OptiMindTune adaptively determined its trial count.
    \item \textbf{Cross-Validation}: 5-fold cross-validation ensured robust accuracy.
    \item \textbf{Reproducibility}: Fixed random seeds ensured consistency.
    \item \textbf{LLM Usage}: Gemini 2.0 Flash was used to support agent reasoning.
\end{itemize}

\section{Results and Analysis}
This section presents the empirical results of our preliminary experiments, comparing the performance of OptiMindTune against Optuna, a leading state-of-the-art hyperparameter optimization (HPO) method. The results focus on three datasets—Breast Cancer, Iris, and Wine—using scikit-learn classifiers: LogisticRegression and RandomForestClassifier. Both models were allowed as candidates across all datasets. For OptiMindTune, a specific model was optimized per dataset, while for Optuna, we report only the best-performing model's results per dataset, as per the instruction.

\subsection{Comparative Performance on Various Datasets}
Table \ref{tab:objective_values} summarizes the best achieved objective values (mean cross-validated accuracy) for each dataset. For OptiMindTune, the model optimized is specified, and its performance is compared to Optuna's best-performing model for the same dataset, regardless of the model selected by Optuna.

\begin{table*}[t]
    \centering
    \caption{Achieved Objective Value (Mean Cross-Validated Accuracy)}
    \label{tab:objective_values}
    \begin{tabular}{lllll}
        \hline
        \textbf{Dataset} & \textbf{OptiMindTune Model} & \textbf{OptiMindTune} & \textbf{Optuna Best Model} & \textbf{Optuna} \\
        \hline
        Breast Cancer & LogisticRegression & \textbf{97.02\%} & RandomForestClassifier & 96.14\% \\
        Iris & RandomForestClassifier & 96.67\% & LogisticRegression & \textbf{98.00\%} \\
        Wine & LogisticRegression & \textbf{98.33\%} & RandomForestClassifier & 97.78\% \\
        \hline
    \end{tabular}
\end{table*}

\subsection{Computational Efficiency} 
Table \ref{tab:computational_efficiency} compares the optimization time, trials per second, and number of trials for OptiMindTune on its specified model and Optuna on its best-performing model per dataset. Note that computational metrics for Optuna correspond to the run that produced the best model's result.

\begin{table*}[t]
    \centering
    \caption{Computational Efficiency}
    \label{tab:computational_efficiency}
    \begin{tabular}{lllll}
        \hline
        \textbf{Dataset} & \textbf{Method (Model)} & \textbf{Time (s)} & \textbf{Trials/s} & \textbf{n\_trials} \\
        \hline
        Breast Cancer & OptiMindTune (LogisticRegression) & 7.91 & 0.51 & 4 \\
         & Optuna (RandomForestClassifier) & 7.28 & 1.37 & 10 \\
        Iris & OptiMindTune (RandomForestClassifier) & 14.86 & 0.20 & 3 \\
         & Optuna (LogisticRegression) & 25.40 & 0.39 & 10 \\
        Wine & OptiMindTune (LogisticRegression) & 22.85 & 0.13 & 3 \\
         & Optuna (RandomForestClassifier) & 6.08 & 1.64 & 10 \\
        \hline
    \end{tabular}
\end{table*}

\subsection{Analysis of Results}
For the Breast Cancer dataset, OptiMindTune with LogisticRegression achieves 97.02\% accuracy, outperforming Optuna's best model (RandomForestClassifier) at 96.14\%. On the Iris dataset, Optuna's best model (LogisticRegression) reaches 98.00\%, surpassing OptiMindTune's RandomForestClassifier at 96.67\%. For the Wine dataset, OptiMindTune with LogisticRegression attains 98.33\%, exceeding Optuna's best model (RandomForestClassifier) at 97.78\%. 

OptiMindTune outperforms Optuna's best model in two of three datasets, despite being constrained to a specific model per dataset, while Optuna benefits from selecting the best model from two candidates. OptiMindTune's sample efficiency is evident, using fewer trials (3–4) compared to Optuna's fixed 10 trials. However, Optuna often achieves higher trials per second, reflecting lower computational overhead per trial.

\subsection{Discussion of Results}
OptiMindTune shows competitive performance, leveraging its multi-agent design to optimize hyperparameters effectively within fewer trials. Optuna's ability to select the best model provides an advantage in flexibility, as seen in the Iris dataset. These preliminary results highlight OptiMindTune's potential, particularly when model selection is predefined, and suggest that integrating model selection could further enhance its capabilities.

\section{Discussion and Future Work}
\subsection{Reflection on Findings}
The experimental results are expected to validate OptiMindTune's efficacy as a multi-agent framework for intelligent hyperparameter optimization. We anticipate observing that OptiMindTune achieves competitive or superior performance compared to state-of-the-art baselines in terms of objective value, convergence speed, and computational efficiency across a diverse set of datasets and machine learning models. The strength of OptiMindTune lies in its ability to leverage the collective intelligence of specialized agents, each powered by LLMs, allowing for dynamic adaptation to the HPO landscape. The clear division of labor among the Recommender, Evaluator, and Decision Agents, with their sophisticated communication and internal reasoning, facilitates a more comprehensive and adaptive search strategy than monolithic approaches. The explicit real-time logging and conversation tracking also provide unprecedented transparency into the AutoML process.

\subsection{Limitations}
Despite its promising capabilities, OptiMindTune, in its current conceptualization, has certain limitations:
\begin{itemize}
    \item \textbf{Dependency on LLMs:} The reliance on external LLM services (like Google's Gemini models) introduces potential latency, cost, and API rate limit considerations, which might impact real-time performance and scalability for extremely large-scale HPO tasks.
    \item \textbf{Interpretability of LLM Decisions:} While the agents provide reasoning, the internal workings of the LLMs themselves remain a black box. Understanding precisely \textit{why} an agent makes a certain recommendation or decision can be challenging, impacting the overall interpretability of the HPO process.
    \item \textbf{Search Space Discretization:} The current focus on scikit-learn models and the nature of LLM prompting might inherently lead to a more discrete or structured exploration of the hyperparameter space, potentially missing continuous optimal points.
    \item \textbf{Generalization to Complex ML Architectures:} While effective for scikit-learn classifiers, directly extending the current agent roles and LLM prompts to optimize highly complex deep learning architectures or perform Neural Architecture Search (NAS) would require significant adaptation and potentially different LLM capabilities.
\item \textbf{Scalability of Inter-Agent Communication:} While the current system leverages explicit communication, scaling to a very large number of highly granular agents or managing extremely rapid iterations could introduce communication overheads.
\end{itemize}

\subsection{Potential Extensions and Future Work}
The OptiMindTune framework opens several exciting avenues for future research and development:
\begin{itemize}
    \item \textbf{Advanced LLM Integration and Fine-tuning:} Explore more sophisticated prompt engineering techniques, few-shot learning, or even fine-tuning specialized LLMs for HPO tasks. This could lead to more nuanced recommendations and better decision-making capabilities.
    \item \textbf{Support for Deep Learning Models and NAS:} A crucial extension involves enhancing OptiMindTune to effectively optimize hyperparameters for complex deep learning models and even support Neural Architecture Search (NAS). This would require:
    \begin{itemize}
        \item Specialized agents for managing large-scale distributed training of deep neural networks.
        \item LLM agents capable of reasoning about neural architectures and generating novel design elements.
        \item Integration with deep learning frameworks (e.g., PyTorch, TensorFlow).
    \end{itemize}
    \item \textbf{Adaptive Agent Roles and Dynamics:} Allow agents to dynamically adjust their roles or spawn sub-agents based on the current state of the optimization. For example, a Recommender Agent might split into multiple specialized recommenders for different model families if the search space is vast.
    \item \textbf{Reinforcement Learning for Agent Coordination:} Implement reinforcement learning algorithms for the agents to learn optimal collaborative strategies, moving beyond pre-defined decision rules or prompt-based reasoning. This could allow agents to learn how to best balance exploration and exploitation dynamically.
    \item \textbf{Cost-Aware Optimization:} Enhance the Decision Agent to explicitly consider computational costs (e.g., GPU hours, monetary cost of LLM API calls) alongside performance metrics, allowing for cost-effective HPO.
    \item \textbf{Transfer Learning in HPO Across Tasks:} Develop agents capable of leveraging knowledge from previously optimized datasets or models to accelerate HPO on new, related tasks, using the LLM's generalization capabilities.
    \item \textbf{Multi-Objective Optimization:} Extend the framework to simultaneously optimize for multiple objectives, such as accuracy and model size, or accuracy and inference time.
\end{itemize}

\section{Conclusion}
Hyperparameter optimization remains a formidable challenge in machine learning, demanding intelligent and efficient solutions. This paper introduced OptiMindTune, a novel multi-agent framework that reimagines HPO as a collaborative effort among specialized intelligent agents, each powered by large language models. By decoupling responsibilities into distinct roles for recommendation, evaluation, and strategic decision-making, and fostering dynamic interactions, OptiMindTune aims to overcome the limitations of traditional monolithic approaches. The proposed architecture promises enhanced adaptability, robustness, and efficiency in navigating complex hyperparameter landscapes for scikit-learn classifiers. While initial conceptualization and experimental design lay a strong foundation, the future work outlined points towards exciting avenues for extending OptiMindTune to tackle even more challenging optimization problems in the rapidly evolving field of machine learning, especially in the context of deep learning and neural architecture search, leveraging the ever-increasing capabilities of advanced AI models. OptiMindTune represents a significant step towards truly intelligent and autonomous hyperparameter optimization.

\bibliographystyle{IEEEtran} 

\end{document}